\documentclass[11pt]{article}

\usepackage[utf8]{inputenc}
\usepackage[T1]{fontenc}
\usepackage{amsmath,amssymb,amsthm}
\usepackage{graphicx}
\usepackage{caption}
\usepackage{subcaption}
\usepackage{natbib}
\usepackage{caption}
\usepackage{authblk}
\usepackage{geometry}
\usepackage{newunicodechar}
\newunicodechar{≥}{\geq}
\usepackage{authblk}
\usepackage{fancyhdr}
\usepackage{xcolor}
\usepackage{booktabs}
\usepackage{multicol}

\usepackage[
    colorlinks=true,
    linkcolor=black,
    citecolor=blue,
    urlcolor=blue
]{hyperref}

\title{LLMs Don't Pay for the Jump}
\author[1]{Paras Balani}
\author[2]{Subhrakanta Panda}
\affil[1]{Department of Mathematics and Department of Computer Science, Birla Institute of Technology and Science, Pilani, Hyderabad Campus, Jawahar Nagar, Kapra Mandal, Medchal District, Telangana 500078, India}
\affil[2]{Department of Computer Science, Birla Institute of Technology and Science, Pilani, Hyderabad Campus, Jawahar Nagar, Kapra Mandal, Medchal District, Telangana 500078, India}
\date{}

\begin{document}

\maketitle

\begin{abstract}
\citet{zahavy2026llms} argues that Large Language Models, despite their capabilities in induction and deduction, cannot perform the abductive ``Jump'' that produced Einstein's equivalence principle, and attributes this limitation to the absence of embodied simulation. \citet{zhengxin2026jump} and \citet{farmer2026abduction} question whether embodiment is necessary for abduction, pointing to alternative routes to General Relativity and forms of abduction that require no sensorimotor grounding. Max Planck resolved the blackbody radiation problem in 1900. Planck's move to \(E=h\nu\) required no embodied simulation. It was motivated by a mathematical consequence of classical theory, an infinite predicted energy for a finite measured quantity, that could not be physically accepted. We show that neither induction nor deduction could have produced the postulate and argue that its adoption required a coupling between epistemic error and physical cost. We formalize this distinction through thermodynamic coupling and show that fixed-weight transformer inference lacks such coupling, regardless of model scale. This is consistent with empirical results showing that output entropy remains nearly unchanged across tasks with sharply increasing causal difficulty, even as accuracy falls from \(100\%\) to \(17\%\). We therefore argue that the missing ingredient in machine abduction may lie deeper than embodiment: a system must have some physical mechanism through which epistemic error becomes costly enough to force revision.
\end{abstract}

\begin{multicols}{2}

\section{Introduction and Related Work}
In 1900, Max Planck encountered a failure that classical physics could not explain away. The Rayleigh-Jeans law, derived from the established principles of classical electrodynamics and statistical mechanics, predicted that a blackbody in thermal equilibrium would emit increasingly large amounts of energy as frequency increased, ultimately implying infinite energy at high frequencies. This became known as the ultraviolet catastrophe. The prediction followed directly from the theory's own assumptions and was decisively contradicted by observation. Planck therefore had to alter the underlying picture itself. He proposed that energy exchange occurs in discrete amounts according to $E = h\nu$, introducing an assumption with no place in classical mechanics.

The legitimacy of that leap requires a different account of what scientific discovery involves. A prominent view developed by Schmidhuber (2008) treats scientific discovery as fundamentally a problem of compression: finding the shortest program, rule, or description that accounts for a set of observations \cite{schmidhuber2008compression}. On this view, a scientist observes recurring regularities in data and searches for a more compact principle that captures them. This makes discovery closely related to \emph{Induction}, where a general rule is inferred from particular observations and results. It also fits naturally with recent progress in systems such as AlphaProof, which has demonstrated strong performance on formal mathematical reasoning through reinforcement learning \cite{hubert2026alphaproof}. Deduction begins with an established set of assumptions and determines what follows from them, while induction attempts to infer a general rule from observed regularities. If scientific discovery were adequately described as the combination of these two operations, then the remaining task would appear to be one of scale and capability. Given enough experimental data, mathematical knowledge, and computational power, a sufficiently capable Large Language Model should, in principle, be able to reconstruct the hypotheses that scientists have discovered.

Under this picture, Planck's quantization hypothesis should have been accessible to a system capable of recognizing the failure of the Rayleigh--Jeans law, compressing the observed blackbody spectrum into a simpler description, and deriving the consequences of that description. But Planck introduced a new constraint on what physical processes were allowed to occur, one that was absent from the classical principles that had generated the anomaly in the first place. The problem therefore cannot be reduced to finding a regularity in the data and deriving its consequences. Planck had to propose a new physical possibility that classical theory did not contain. So, can a scientific discovery be reduced to the combination of induction and deduction, or whether some discoveries require changing the space of possible explanations itself.

The Rayleigh-Jeans catastrophe was not a compression problem: there was no dataset to compress, because the observations that would eventually confirm quantized radiation had not yet been collected in a form that pointed to quantization. Nor was it a deduction problem: deduction from classical premises is precisely what \emph{produced} the catastrophe. What Planck required was a third operation - the generation of an axiom that neither the data nor the existing rules could supply, motivated instead by the sheer physical unacceptability of the alternative. Peirce (1934) named this operation \emph{Abduction}: the inference of a Case, or a new Rule, from a Rule and a surprising Result \cite{peirce1934collected}.

\begin{itemize}
    \item \textbf{Deduction} (Rule + Case $\rightarrow$ Result): the analytic application of a rule to a case, the only mode of inference that guarantees truth.
    \item \textbf{Induction} (Case + Result $\rightarrow$ Rule): the synthetic extraction of a general rule from repeated cases and results, validated by statistical frequency.
    \item \textbf{Abduction} (Rule + Result $\rightarrow$ Case): the inference of a case, or a new rule, that would explain an otherwise inexplicable result.
\end{itemize}

Unlike deduction, abduction does not guarantee truth, and unlike induction, it does not require repeated cases. It responds to a result that a system cannot afford to leave unexplained. We take ``cannot afford'' literally. Planck did not reject classical electrodynamics because of accumulated statistical evidence, but because its prediction of infinite radiated energy at finite temperature was physically untenable. The distinction, we argue, lies in the cost of error. Small discrepancies can be absorbed through corrections or revised assumptions, but an unbounded contradiction demands new axioms. Genuine abduction begins when a theory can no longer afford to preserve its existing rules.

\citet{zahavy2026llms}, in \emph{LLMs Can't Jump}, makes a similar argument using Einstein's formulation of General Relativity. He argues that Newtonian physics faced no decisive empirical crisis when Einstein began his work, so Einstein's abductive leap, expressed through the equivalence principle, did not emerge from data compression. Instead, Zahavy attributes it to embodied simulation, particularly Einstein's imagined experience of an observer in a falling elevator. Drawing on \citet{magnani2009abductive}'s account of \emph{manipulative abduction} and \citet{harnad1990symbol}'s symbol grounding problem, Zahavy argues that LLMs manipulate physical language without access to the sensory and embodied experience that can give physical concepts meaning. He therefore proposes physically consistent, multimodal, action-controllable world models, such as those explored by \citet{bruce2024genie} in Genie, as a possible route toward such grounding.

But if embodiment is the missing ingredient in scientific abduction, then the problem is fundamentally one of modality. LLMs lack direct sensory access and physical interaction, so they cannot ground their representations through perception or action. Planck's case does not fit this explanation. His abductive move involved no bodily experience or physical interaction. He encountered a mathematical consequence of classical statistical mechanics: the Rayleigh-Jeans law implied an unbounded amount of radiated energy at high frequencies. Planck responded by changing the formal assumptions governing the distribution of energy among electromagnetic modes and introducing the relation $E = h \nu$. The source of the new hypothesis was a contradiction within the mathematical theory itself. No sensory experience was required to formulate it.

\citet{zhengxin2026jump}, in a public reply to Zahavy, raises a similar objection from another direction. He notes that Einstein's route to General Relativity was not the only possible one. Feynman's later reconstruction begins from special relativity and quantum field theory, models gravity through a massless spin-2 graviton, and reaches the same field equations through a largely deductive route. 

\citet{farmer2026abduction} on the other hand, argues that while most scientific abduction may require sensorimotor coupling, an exception is ``identity abduction,'' where two independently developed structures are recognized as equivalent through a shared representation. This can occur without physical interaction when a diagram or other representation reveals an invariant connecting them. Einstein's equivalence principle may fit this category: the recognition that inertial and gravitational mass are the same quantity. This raises the possibility that Einstein's insight depended less on embodiment itself than on a representational relation revealed through an embodied thought experiment. Taken together, these objections point to the same gap in the embodiment account. Embodiment may provide a sufficient mechanism for abduction, as Einstein's elevator thought experiment suggests, but it does not appear to be necessary in every case. Planck's quantization hypothesis provides a historical example in which the abductive move arose from a formal contradiction without physical interaction or sensory experience. 

A related concern appears in the work of \citet{floridi2025reasoning}, who approach the problem from the philosophy of information. They argue that LLMs have a ``stochastic core'' and only an ``abductive appearance.'' On their account, an LLM can produce statements that resemble hypotheses generated through inference to the best explanation, but this apparent reasoning may arise from statistical learning over human-generated text. The training corpus already contains the results of scientific reasoning, including theories, explanations, arguments, and descriptions of evidence. The model can therefore reproduce combinations of these products without performing the original inferential process that produced them. So basically, successful hypothesis generation alone does not establish that the system has performed abduction. A system may produce a plausible explanation by exploiting patterns in a corpus that already contains explanations, without independently identifying a surprising result and constructing a new hypothesis to account for it.

\citet{floridi2025categorical} develop a related argument using category theory and the problem of symbol grounding. Their claim is that LLMs do not necessarily solve the grounding problem by acquiring direct connections between symbols and the physical world. Instead, they can operate on symbols that have already been grounded by human agents. This connects their argument to Harnad's account of epistemic parasitism. Human scientists acquire concepts through interaction with the world and then encode the resulting knowledge in language, mathematics, diagrams, and other symbolic systems. An LLM can subsequently learn from these symbolic products without reproducing the original process through which those symbols acquired their physical meaning. The model therefore inherits a grounded conceptual structure from the corpus without having to establish the grounding itself.

\citet{sun2025occam} introduce InAbHyD, a synthetic benchmark designed to test whether LLMs can generate abductive hypotheses that are both correct and parsimonious, favoring the simplest explanation consistent with the evidence. They find that models perform reasonably well in simple world models, but their performance declines sharply as the underlying world model becomes more complex. This degradation persists despite in-context learning and reinforcement learning from verifiable rewards. \citet{salimi2026wiring} provide the first comprehensive survey of abductive reasoning in LLMs. They formalize abduction as a two-stage process consisting of Hypothesis Generation and Hypothesis Selection, and review more than sixty studies across eleven model families ranging from 3B to 72B parameters. Their survey identifies a persistent gap in the field: limited mechanistic understanding of how abductive reasoning occurs. Existing work can show that abduction becomes unreliable as problems increase in complexity and novelty, but it has not yet established what properties a computational system must possess for abduction to succeed.

This is the question we take up. Zahavy asks what cognitive mechanism Einstein possessed that LLMs lack, while Zheng-Xin and Farmer question whether such a mechanism must be embodied at all. We ask what lies beneath these explanations: what must be true of the physical substrate of any bounded predictive system for an error to become costly enough to force the abandonment of a rule? This motivates our choice of Planck rather than Einstein as the central case study. Planck's abduction was purely formal, and the crisis he faced had an unusually clear structure: classical theory predicted that a finite, measurable quantity must be infinite. His problem therefore makes the cost of persisting with an erroneous rule explicit. We argue that this cost is more than a metaphor. Any physical computation, including inference in a language model, is subject to thermodynamic constraints. \citet{landauer1961irreversibility} showed that logically irreversible computation has an associated physical cost, with the erasure of one bit requiring a minimum dissipation of $k_{\mathrm{B}}T\ln 2$ under the standard Landauer bound. This sets a physical limit on computation, but it does not determine how costly it is for a system to maintain a false model of the world. That second quantity is logically independent of computational cost.

A system may have sufficient resources to compute an answer while having no internal pressure to abandon the rule that produced it. This distinction also changes how we interpret the role of embodiment. Zahavy identifies the absence of embodiment as a possible obstacle to abduction because sensory and physical interaction may provide the grounding needed to construct new hypotheses. Floridi and his coauthors raise a different concern: an LLM may operate on representations that were already grounded by human agents, so adding sensory inputs would not by itself explain how the system generates or grounds its own concepts. Our argument shifts the focus from the modality through which a system represents the world to the physical consequences of maintaining an erroneous representation.

\section{Background}

\subsection{Crisis of Classical Radiation}
Why does a physical anomaly sometimes force scientists to abandon a theory, while in other cases the theory survives after being modified? The answer depends on where the failure occurs. A physical theory connects assumptions about how nature works to mathematical equations that generate predictions. When an observation disagrees with a prediction, the disagreement does not automatically show that the theory is false. The problem may come from an inaccurate approximation, an incomplete calculation, or an assumption that applies only within a limited range of conditions. The situation becomes much more serious when the prediction follows directly from the theory's basic assumptions and the prediction is clearly contradicted by reliable observations. This was the problem Max Planck encountered in 1900 with blackbody radiation. A blackbody is an idealized object that absorbs radiation and reaches thermal equilibrium with its surroundings. Physicists wanted to calculate how much electromagnetic energy such an object should emit at each frequency. Classical physics provided the necessary tools: classical electrodynamics described electromagnetic radiation, while statistical mechanics described how energy should be distributed among the possible modes of the electromagnetic field. From these principles, the Rayleigh--Jeans law was derived. At low frequencies, the law agreed with observations. At high frequencies, however, it predicted that the emitted energy would increase without limit. If the law were correct at all frequencies, the total energy emitted by a blackbody would be infinite. This was called the ultraviolet catastrophe because the problem became apparent at ultraviolet frequencies. The important point is that the divergence was not caused by a faulty measurement or a lack of data. It came from the mathematical consequences of classical theory itself. Keeping the classical assumptions unchanged meant accepting a prediction that physical systems plainly did not obey.

Planck responded by changing the assumption about how matter exchanges energy with radiation. Classical physics treated energy as continuous, so an oscillator could gain or lose any amount of energy. Planck proposed that energy exchange occurred in discrete units, with the energy of each unit given by
\[
E = h\nu.
\]
Here, \(E\) is the energy of one unit, \(\nu\) is the frequency of the radiation, and \(h\) is a new constant of nature, now called Planck's constant. This proposal introduced a new idea into physics. An oscillator associated with a high frequency could no longer exchange an arbitrarily small amount of energy. Each exchange required an amount proportional to that frequency. At high frequencies, the required energy therefore became large compared with the thermal energy available at a fixed temperature. High-frequency modes became much less likely to be excited, which prevented the unlimited accumulation of energy predicted by the Rayleigh--Jeans law. Planck's formula consequently reproduced the observed blackbody spectrum.

\subsection{Boltzmann, Entropy, and the Cost of Being Wrong}
Planck's route to \(E=h\nu\) did not begin with radiation. It began with an idea he borrowed, reluctantly, from Ludwig Boltzmann, a physicist whose statistical approach remained controversial among many of his contemporaries. In the 1870s, Boltzmann proposed that entropy could be understood in terms of the number of microscopic arrangements compatible with a system's observed macroscopic state. In its familiar form, this relation is
\[
S = k_B \ln W,
\]
where \(S\) is entropy, \(k_B\) is Boltzmann's constant, and \(W\) denotes the number of possible microscopic configurations corresponding to the same macroscopic state \citep{boltzmann1877,planck1901}. So, this connected a thermodynamic quantity, entropy, to the hidden microscopic structure of matter. To use \(W\) in this way, however, one had to assume that matter consisted of discrete atoms and molecules whose possible arrangements could be counted, even though those microscopic constituents could not be directly observed.

 Physicists such as Ernst Mach and Wilhelm Ostwald rejected the idea that atoms should be treated as physically real entities. They regarded atoms as useful theoretical constructions that could organize observations without committing physics to an unseen microscopic reality. From this perspective, thermodynamics should be formulated using quantities that could be measured directly, such as energy, temperature, and entropy. Boltzmann's statistical mechanics took a different position. It treated macroscopic thermodynamic behavior as the result of an enormous number of microscopic states, even though those states were inaccessible to direct observation. Planck initially shared many of the reservations about Boltzmann's approach, which makes his later use of Boltzmann's statistical reasoning important. To derive his radiation law, Planck would have to rely on a microscopic picture that he had not previously regarded as physically secure \citep{janssen2019planck,flam1997boltzmann}.

Boltzmann maintained his position despite this opposition, and the cost was personal as well as professional. His statistical approach faced sustained criticism from influential physicists, while the existence of atoms remained disputed throughout much of his career. He died in 1906, before Jean Perrin's experiments on Brownian motion provided strong experimental support for atomism, building on Einstein's theoretical analysis of Brownian motion in 1905. The history of Boltzmann's later years has therefore often been discussed in connection with the isolation and despair he experienced while defending a microscopic view of matter that many of his contemporaries rejected.

Boltzmann's statistical framework provided the mathematical machinery that made Planck's abduction possible. Planck used
\[
S = k_B \ln W
\]
to count the ways a fixed energy could be distributed among discrete oscillators. This counting connected Planck's quantization postulate to entropy and led to the specific relation
\[
E = h\nu.
\]
Boltzmann's method therefore supplied the formal bridge between Planck's new assumption and the radiation law it produced.

But there is a second, more structural reason Boltzmann belongs in this story, and it is the reason we place him between the crisis (Section~2.1) and its resolution (Section~2.3) rather than treating him as a mere technical aside. His atomism shows that defending a rejected axiom can carry real professional and personal costs. An axiom that survives sustained resistance therefore differs from one that is proposed casually and abandoned when difficulties arise. We return to this asymmetry in Section~5, where we argue that the cost of maintaining a commitment, rather than the accumulation of confirming evidence alone, helps distinguish genuine abduction from an ordinary, revisable guess.

\subsection{Planck's Quantization}
Planck did not arrive at \(E=h\nu\) because he found Boltzmann's statistical mechanics convincing. By the autumn of 1900, he had exhausted the classical approaches he trusted, and he later described his decision as ``an act of desperation,'' undertaken because ``a theoretical interpretation had to be found at any cost.'' Throughout the 1890s, he had tried to derive the blackbody spectrum within classical electrodynamics and thermodynamics. When these approaches failed, he turned to Boltzmann's statistical counting, despite his earlier reservations about the underlying microscopic picture.

So basically, Planck introduced discrete energy elements because the continuous alternatives available within classical theory all led back to the same divergence. Even after introducing \(E=h\nu\), he initially treated quantization as a formal device for obtaining the correct radiation law. Its later interpretation as a physical principle developed through Einstein's 1905 work on the photoelectric effect and subsequent quantum theory. Planck's move therefore fits Peirce's account of abduction: \(E=h\nu\) was neither induced from an established pattern nor deduced from existing principles. It was a new axiom introduced because preserving the existing rule system led to a physically unacceptable result.

\section{The Limits of Inductive Inference}
Compression rewards a rule to the extent that the rule makes existing data more predictable; it offers no comparable reward for a rule that also happens to prevent a \emph{future}, not-yet-collected measurement from contradicting a \emph{different} theory's extrapolation. Nothing about a compression objective, evaluated against the data available in 1900, would have penalized the Rayleigh--Jeans law for its behavior at frequencies where no data yet existed to compress, since a compression-driven search only optimizes over the observations it has been given. Planck's abduction was motivated by exactly this kind of extrapolated, not-yet-observed failure: the mathematical certainty that the classical law, followed to its logical conclusion, would demand an infinite quantity. No accumulation of the low- and mid-frequency data actually in hand, however large, could have surfaced that failure through pattern-matching alone. What was needed instead was a system able to take the classical rule's own unconfirmed extrapolation seriously enough to treat its consequence as intolerable before that consequence had been measured.

As Section~2.3 notes, the experiments that later established quantization as a physical principle, including Einstein's 1905 explanation of the photoelectric effect, Bohr's 1913 atomic model, and Compton's 1923 scattering results, were not yet available in 1900. Induction can infer only from the cases and results available to it. It cannot directly anticipate a rule whose strongest confirmation will come from future observations. By the standard of fit to existing data, Planck's interpolation formula was already sufficient. The deeper claim \(E=h\nu\) was not demanded by the evidence available at the time. Its significance came from making a specific, falsifiable claim about microscopic energy exchange before the evidence needed to support that claim had appeared.

Compression therefore provides no natural incentive to prefer a rule because it avoids a future contradiction that is absent from the current dataset. A compression objective evaluates how well a rule explains the observations it has already been given. It has no direct penalty for the behavior of a theory in an unobserved regime.

\section{The Limits of Deduction}
Once Planck's postulate was in place, the problem changed from proposing a new principle to deriving its consequences. Given \(E=h\nu\) as an axiom, the derivation of the blackbody spectrum, the recovery of the Rayleigh--Jeans law in the low-frequency limit, and the Stefan--Boltzmann law from the total emitted energy are deductive tasks. In Peirce's schema, they follow the form Rule + Case $\rightarrow$ Result: once the premises are fixed, the task is to determine what follows from them. This is the class of reasoning at which modern formal systems perform well. Proof assistants based on dependent type theory and reinforcement-learned systems such as AlphaProof \citep{hubert2026alphaproof} can derive complex results from explicitly stated premises, with recent systems reaching strong performance on formal mathematical benchmarks.

Deduction, however, cannot produce the axiom from which the deduction begins. An axiom is a premise, not a theorem derived from prior premises. No chain of valid deductions from classical electrodynamics and classical statistical mechanics could yield \(E=h\nu\), because those premises treat energy exchange as continuous. The Rayleigh--Jeans catastrophe was therefore not a failure of deduction. It was the correct deductive consequence of the classical assumptions available in 1900. A system restricted to those assumptions could derive the divergent energy integral with complete logical accuracy, but it could not use deduction alone to replace the assumptions that produced it. Deduction explores the logical space defined by a set of premises. Planck's move required leaving that space and introducing a premise that the existing theory did not contain. This is the distinction between deriving the consequences of a theory and generating a new theory.

Suppose a modern reasoning system were asked to find inconsistencies in nineteenth-century physics. It could identify the Rayleigh--Jeans divergence as a mathematically anomalous result: the theory predicts a divergent integral for a quantity that experiments show to be finite. A sufficiently thorough search could therefore flag the catastrophe as an anomaly. But detecting an inconsistency does not determine whether it requires a change in theory. Classical physics contained many approximations and idealizations that were tolerated without prompting a revision of its foundations. The difference was the cost of the error. A \(2\%\) discrepancy might be attributed to measurement error or an imperfect approximation, whereas an infinite prediction for a finite observable quantity cannot be absorbed in the same way. Recognizing that distinction requires more than deduction. The system must have some prior criterion for how costly an error is and when that cost becomes unacceptable. Without such a criterion, a deductive search can identify anomalies but cannot determine which ones justify abandoning an existing rule. Planck's own decade of unsuccessful attempts to preserve the classical account illustrates this distinction: the problem was visible before he accepted that the cost of retaining the classical framework had become too high.

\section{Thermodynamic (De)coupling: Formal Framework}

\subsection{Descriptive versus Regulatory Uncertainty}
We distinguish two roles that uncertainty can play in a predictive system, following the terminology of \citet{gamaleldin2026descriptive}.

\textbf{Definition 1 (Thermodynamic coupling).} A computational system is \emph{thermodynamically coupled} if the physical cost of an operation, \(E_{\mathrm{cost}}\), increases with the epistemic error of its output, \(\varepsilon = |\hat{y}-y^*|\):
\[
\frac{\partial E_{\mathrm{cost}}}{\partial \varepsilon} > 0.
\]

In a coupled system, epistemic error has a direct physical consequence. Maintaining an incorrect prediction becomes more costly as the error increases, creating a physical pressure to revise the prediction or the rule that produced it. Uncertainty is therefore \emph{regulatory}: the system's physical cost depends on the accuracy of its predictions, so persistent error creates pressure for correction.

\textbf{Definition 2 (Thermodynamic decoupling).} A system is \emph{thermodynamically decoupled} if
\[
\frac{\partial E_{\mathrm{cost}}}{\partial \varepsilon}=0,
\]
so that the energetic cost of an operation is independent of its epistemic error, conditional on the state of the physical substrate. Uncertainty in such a system is therefore \emph{descriptive}: it characterizes the system's predictive uncertainty without affecting the physical cost or subsequent dynamics of the computation. Whether a prediction is correct or incorrect does not, by itself, change the energetic cost of producing it.

Every operation performed by a bounded physical system remains subject to Landauer's (1961) bound, with each irreversible bit erasure dissipating at least \(k_B T \ln 2\) joules. This cost applies regardless of whether the erased information corresponds to a correct inference or an erroneous prediction. The distinction we draw is narrower: whether the physical cost of computation depends on epistemic quality, or whether it remains independent of whether the system is correct.

This distinction allows us to restate the limits identified in Sections~3 and~4. Planck's crisis was difficult to ignore because the classical theory produced an unbounded physical consequence. An infinite predicted energy could not be treated as a small residual or absorbed through a minor correction. Planck spent years attempting to preserve the classical framework before the cost of maintaining it became greater than the cost of abandoning its underlying assumptions. Boltzmann's history shows the same asymmetry at the level of a scientific career. Defending atomism imposed sustained professional and personal costs despite limited acceptance at the time. In both cases, the pressure to change a commitment came from the cost of maintaining it, not simply from the accumulation of evidence against it. We argue that this distinction separates genuine abductive commitment from an ordinary hypothesis that can be revised without significant cost.

\subsection{The Softmax--Gibbs Analogy}
The softmax function used to convert logits into a token distribution,
\[
P(v_i \mid \mathrm{ctx}) =
\frac{\exp(x_i/\tau)}
{\sum_j \exp(x_j/\tau)},
\]
has the same mathematical form as the Gibbs--Boltzmann distribution over microstates, with the inference-time parameter \(\tau\) playing a formal role analogous to temperature.

\textbf{Theorem 1 (Softmax decoupling).} In a transformer with fixed weights \(\theta\) and fixed inference temperature \(\tau\), the token-level Shannon entropy
\[
H_t =
-\sum_i P_\theta(v_i \mid \mathrm{ctx}_t)
\log P_\theta(v_i \mid \mathrm{ctx}_t)
\]
is a deterministic function of \((\theta,\tau,\mathrm{ctx}_t)\) alone. It is therefore independent of whether \(\mathrm{ctx}_t\) is in-distribution or requires causal extrapolation beyond the training data.

For fixed \(\theta\) and \(\mathrm{ctx}_t\), the transformer produces a fixed logit vector \(x_t\). The temperature \(\tau\) determines the corresponding token probabilities through the softmax function, and these probabilities uniquely determine \(H_t\). No variable in this computation represents the physical temperature of the hardware, the instantaneous power dissipated by the computation, or the eventual correctness of the generated token. Consequently, two contexts can produce the same predictive entropy while differing in their epistemic status. One may correspond to a familiar pattern that the model has learned reliably, while another may require an extrapolation unsupported by the training distribution. The softmax entropy records the distribution of the model's output probabilities, but the computation that produces those probabilities does not assign any additional physical cost according to whether the prediction is correct or incorrect.

\subsection{Cost-Coupled Abduction: A Proposed Criterion}
We can now state the criterion developed in the preceding sections. Induction and deduction can identify patterns and derive consequences, but neither provides a mechanism for treating a particular error as intolerable. Section~5.1 called this missing mechanism regulatory uncertainty, while Section~5.2 showed that transformer inference exhibits the opposite structure: token entropy describes the output distribution without depending on whether the resulting claim is correct.

\textbf{Criterion (Cost-Coupled Abduction).} A bounded predictive system can generate a genuinely new axiom only if some internal cost is causally coupled to the magnitude of its epistemic error on a specific result. Formally,
\[
\frac{\partial E_{\mathrm{cost}}}{\partial \varepsilon}>0,
\]
where the coupling must operate locally when a particular anomaly is encountered, so that maintaining the rule responsible for the error becomes increasingly costly until the rule is revised.

Planck's history is consistent with this criterion, although it does not provide a direct measurement of his cognitive cost function. The relevant proxies are the singular divergence he could not absorb as an ordinary approximation, his years of unsuccessful attempts to preserve the classical framework, and the costs associated with maintaining controversial commitments in Boltzmann's case. These pressures preceded the later evidence that confirmed quantization, so the commitment cannot be explained solely by inductive reward from subsequent observations.

Current LLM inference does not exhibit this form of coupling. A correct and incorrect output produced by the same model on the same hardware does not incur a systematically different computational cost, and Theorem~1 shows that output entropy is determined by the model and context rather than by the truth of the resulting claim. An LLM can reproduce Planck's derivation once \(E=h\nu\) is supplied as a premise, but nothing in the inference process makes the absence of that premise increasingly costly when the classical theory produces an anomaly. The proposed missing ingredient is therefore not deductive ability or sensory modality alone, but a physical mechanism through which epistemic error creates a local cost that can force revision of the rule producing it.

\section{Empirical Signatures of Decoupling}

\citet{gamaleldin2026descriptive} tested whether token-level Shannon entropy tracks causal difficulty using tasks that ranged from retrieval of established results (``Kepler''), to causal reasoning with novel parameters (``Newton''), to extrapolation beyond plausible training coverage (``Newton OOD''). Across three Llama models spanning two orders of magnitude in size, from 3B to 70B parameters, entropy remained statistically flat within each model across the three categories, with ranges of only \(0.011\) to \(0.028\) nats and all \(p \geq 0.568\). Accuracy, however, ranged from \(0\%\) to \(100\%\). The 70B model achieved perfect accuracy on Kepler tasks but only \(17\%\) on Newton OOD, while entropy differed by just \(0.011\) nats between the two. The same pattern appeared in \(N=1{,}000\) examples evaluated on GPT-4o-mini with programmatic scoring: accuracy varied by \(33.2\) percentage points while entropy varied by only \(0.019\) nats.

These results are consistent with the decoupling described by Theorem~1. The model's predictive entropy changes little as the task shifts from familiar retrieval to difficult causal extrapolation, even when accuracy changes substantially. Larger models also become more confident overall, without a corresponding increase in the relationship between confidence and correctness.

\citet{sun2025occam} find that LLMs generate high-quality, parsimonious hypotheses in simple scenarios but struggle as the underlying world models become more complex. This degradation persists under both in-context learning and reinforcement learning from verifiable rewards. \citet{salimi2026wiring}, synthesizing more than sixty studies across eleven model families, report a similar pattern and identify ``limited mechanistic understanding of abductive processes'' as an open problem. Taken together with the entropy results, these findings point to a more specific form of insensitivity: thermodynamic decoupling, where increasing epistemic difficulty does not produce a corresponding physical cost that could drive the system toward revision.

\section{Toward Thermodynamically Coupled Architectures}
We are not proposing thermodynamic coupling as a wholly new requirement. Biological nervous systems provide an existing example of such coupling, most prominently formalized by Friston's (2010) free-energy principle \citep{friston2010freeenergy}. In this framework, the nervous system minimizes a bound on surprise through prediction error and its estimated precision. Prediction error therefore has consequences for the system's physical state: unresolved errors can drive changes in synaptic activity, attention, autonomic responses, and metabolic demand. In the terminology of Section~5.1, biological prediction error is regulatory because epistemic error is linked to the physical processes that maintain and update the system.

This provides a concrete comparison with Planck's case and with transformer inference. If a predictive system is genuinely thermodynamically coupled, larger or more persistent errors should produce measurable changes in physiological or computational resource allocation, such as increased arousal, sustained attention, or reallocation of processing resources. Theorem~1 shows that current transformer inference has no corresponding coupling in its token-level uncertainty. This comparison does not establish that biological systems can perform abduction while transformers cannot. It instead identifies the missing property of a thermodynamically decoupled system: epistemic error has no physical consequence that would compel the system to revise the rule producing it.

One possible engineering direction is neuromorphic hardware in which prediction error continues to drive weight updates during inference, with the energy cost of each update increasing with the magnitude of the error being corrected. This corresponds to the ``Intrinsic Cost'' mechanism proposed by Gamal Eldin (2025). Such a system would couple computational cost to epistemic error: sustaining an incorrect prediction would consume additional energy until the underlying representation or rule was revised. This would create a physical analogue of the costs associated with Boltzmann's defense of atomism and Planck's repeated attempts to preserve the classical account.

\section{Limitations}

A framework that explains past observations is not sufficient unless it also makes predictions that could fail. Section~5.3 therefore yields two direct tests. First, if a thermodynamically coupled architecture of the kind proposed in Section~7 is evaluated on the task suite from Section~6, its internal cost or uncertainty signal should track accuracy as causal demand increases, unlike the entropy of current transformers. If the signal remains flat despite declining accuracy, this would count as evidence against the criterion itself.

Second, the criterion predicts an effect on hypothesis quality, not only calibration. If the degradation in abductive performance reported by \citet{sun2025occam} reflects thermodynamic decoupling, then a coupled system should show less degradation as world-model complexity increases. If hypothesis quality declines by the same amount while uncertainty becomes better calibrated, coupling would improve calibration without improving abduction, weakening the stronger form of the criterion. We do not test these predictions here.

\section{Conclusion}
\citet{zahavy2026llms} approaches the problem through Einstein: what allows a scientist to move from an existing theory to the axioms of a new one when neither accumulated data nor deduction can provide those axioms? We approach the same question through Planck. His route to \(E=h\nu\) required no embodied thought experiment. It began with a finite, measured quantity that classical theory predicted to be infinite \citep{planck1901}. Sections~3 and~4 showed why neither induction nor deduction could supply the required postulate. The available data could already be fitted without quantization, while the classical premises produced the catastrophe when their consequences were derived correctly. What was missing was a mechanism that made the anomaly costly enough to demand a change in the underlying rule. Boltzmann's experience defending atomism provides a second example of such sustained cost.

Section~5 formalized this idea as thermodynamic coupling. A system is coupled when the physical cost of computation depends on epistemic error, and decoupled when that cost is independent of correctness. Theorem~1 shows that token entropy in a fixed transformer does not provide such coupling. Section~6 then showed empirically that entropy remains nearly unchanged even as accuracy falls sharply across tasks and model scales. Section~7 considered possible coupled systems, drawing on biological prediction-error dynamics and neuromorphic architectures. Section~8 turned this proposal into an empirical question: whether introducing such coupling would improve abductive performance under increasing causal and model complexity remains to be tested.

The scope of the argument is deliberately limited. Our case comes from the physical sciences, where theoretical error can produce a measurable physical consequence and where cost therefore has a concrete referent. We do not claim that the same criterion applies unchanged to mathematics or computer science, where the objects of inquiry are formal systems. Within the physical sciences, our claim is narrower: current language models may possess substantial capabilities for induction and deduction, yet their computational substrate contains no mechanism that makes epistemic error itself increasingly costly. Cost-coupled abduction proposes that such a mechanism may be necessary when a system must decide that an existing rule has become too expensive to retain.
\end{multicols}
\bibliographystyle{plainnat}
\bibliography{references}

\appendix

\section{Author's Note}
I am pursuing an MSc in Mathematics alongside a B.E. in Computer Science, while working on cosmology problems involving Einstein's field equations in modified theories of gravity. My work is largely mathematical and computational, which has made the role of assumptions in physical theories particularly concrete. In modified gravity, changing parameters of an imposed energy condition can lead to a different set of field equations and, consequently, a different physical theory. This raises a question that deduction alone cannot answer: why choose one consistent set of assumptions over another, and what would it mean for an AI system to generate such assumptions?

Zahavy's (2026) work gave me a vocabulary for this question through Peirce's triad and the role of abduction. Reading it alongside Gamal Eldin's (2026) work on thermodynamic decoupling led to the synthesis developed in this paper. My mathematical and computational background made the entropy-flatness results particularly natural to interpret, while my work with modified gravity made the problem of generating new axioms feel closely connected to problems I encounter in theoretical physics. It is important to note that this work does not attempt a novel historical or physical account of Planck's route to quantization, since such analysis falls outside my expertise. Instead, I draw on established sources to examine the conditions under which AI systems might move beyond inference toward genuine scientific discovery.

\section{Supplementary Formal Details}
\begin{enumerate}
    \item \textbf{Classical Equipartition:} At thermal equilibrium, classical statistical mechanics assigns each electromagnetic mode an average energy \(k_B T\), independent of frequency. Combined with the classical mode density \(8\pi\nu^2/c^3\), this yields the Rayleigh--Jeans law and the ultraviolet catastrophe: the number of high-frequency modes grows without bound while each receives the same energy.

    \item \textbf{Boltzmann's Statistical Entropy:} Entropy is related to the number of microscopic configurations \(W\) compatible with a macroscopic state,
    \[
    S = k_B \ln W.
    \]
    As discussed in Section~2.2, this required treating matter as composed of discrete, countable microstates, a controversial assumption defended by Boltzmann. This statistical framework supplied the machinery Planck needed to formulate his quantization.

    \item \textbf{Planck's Quantization:} Planck proposed that energy exchange occurs in discrete units,
    \[
    E = h\nu.
    \]
    This postulate was not derived from the previous assumptions. Planck introduced it because maintaining continuous energy exchange alongside the observed finite blackbody spectrum was untenable. Combined with Boltzmann's counting, it gives
    \[
    \langle E(\nu,T)\rangle
    =
    \frac{h\nu}{\exp\left(h\nu/k_B T\right)-1},
    \]
    which approaches \(k_B T\) when \(h\nu \ll k_B T\) and suppresses high-frequency contributions exponentially.

    \item \textbf{The Correspondence Requirement:} Any replacement for classical equipartition must reduce to it in the regime where the classical law was already correct. Item 3's average energy satisfies
\[
\lim_{h\nu/k_B T \to 0} \langle E(\nu,T)\rangle = k_B T,
\]
recovering Item 1 at low frequency or high temperature, exactly where the Rayleigh--Jeans law matched observation. A postulate that failed this limit would not have been an admissible replacement, regardless of how well it resolved the divergence. This is the same constraint discussed in Section~2.3, where Planck initially treated quantization as a formal device rather than a physical claim.

\item \textbf{Thermodynamic Decoupling (Formal Statement):} For a bounded computational system with fixed weights \(\theta\), fixed inference temperature \(\tau\), and inference-time output entropy \(H_t\) as defined in Theorem~1, decoupling holds when
\[
\frac{\partial E_{\mathrm{cost}}}{\partial \varepsilon}=0,
\qquad
\varepsilon = |\hat{y}-y^*|,
\]
conditioned on the substrate state \((T,P)\). Proposition~B.1 and Theorem~1 (Section~5.2) establish that \(H_t\) is a function of \((\theta,\tau,\mathrm{ctx}_t)\) alone, with no dependence on \(\varepsilon\), \(T\), or \(P\). This item states the general decoupling condition, of which Theorem~1's entropy result is one specific instance, restricted to token-level Shannon entropy as the observable proxy for uncertainty.
\end{enumerate}

\end{document}